\documentclass{article} % For LaTeX2e
\usepackage{iclr2021_conference,times}
\usepackage{amsmath,amsfonts,bm}

\def\eqref#1{equation~\ref{#1}}
\def\1{\bm{1}}

\DeclareMathAlphabet{\mathsfit}{\encodingdefault}{\sfdefault}{m}{sl}
\SetMathAlphabet{\mathsfit}{bold}{\encodingdefault}{\sfdefault}{bx}{n}

\usepackage{hyperref}
\usepackage{url}
\usepackage{amsthm,amsmath,amssymb,amsfonts}
\usepackage{color}
\usepackage{graphicx}
\usepackage{booktabs}
\usepackage{algorithm}\usepackage{algpseudocode}
\usepackage{multicol,multirow}
\usepackage{array}
\newcommand{\PreserveBackslash}[1]{\let\temp=\\#1\let\\=\temp}
\newcolumntype{C}[1]{>{\PreserveBackslash\centering}p{#1}}
\newcolumntype{R}[1]{>{\PreserveBackslash\raggedleft}p{#1}}
\newcolumntype{L}[1]{>{\PreserveBackslash\raggedright}p{#1}}

\newtheorem{defn}{Definition}[section]

\newtheorem{assumption}{Assumption}[section]

\newcommand{\tabincell}[2]{\begin{tabular}{@{}#1@{}}#2\end{tabular}}

\makeatletter

\newcommand{\Rmnum}[1]{\expandafter\@slowromancap\romannumeral #1@}
\makeatother

\title{EM-RBR: a reinforced framework for knowledge graph completion from reasoning perspective}

\author{Antiquus S.~Hippocampus, Natalia Cerebro \& Amelie P. Amygdale \thanks{ Use footnote for providing further information
about author (webpage, alternative address)---\emph{not} for acknowledging
funding agencies.  Funding acknowledgements go at the end of the paper.} \\
Department of Computer Science\\
Cranberry-Lemon University\\
Pittsburgh, PA 15213, USA \\
\texttt{\{hippo,brain,jen\}@cs.cranberry-lemon.edu} \\
\And
Ji Q. Ren \& Yevgeny LeNet \\
Department of Computational Neuroscience \\
University of the Witwatersrand \\
Joburg, South Africa \\
\texttt{\{robot,net\}@wits.ac.za} \\
\AND
Coauthor \\
Affiliation \\
Address \\
\texttt{email}
}

\begin{document}

\maketitle
\begin{abstract}
Knowledge graph completion aims to predict the new links in given entities among the knowledge graph (KG). Most mainstream embedding methods focus on fact triplets contained in the given KG, however, ignoring the rich background information provided by logic rules driven from knowledge base implicitly. To solve this problem, in this paper, we propose a general framework, named EM-RBR(\emph{\underline{em}bedding and \underline{r}ule-\underline{b}ased \underline{r}easoning}), capable of combining the advantages of reasoning based on rules and the state-of-the-art models of embedding. EM-RBR aims to utilize relational background knowledge contained in rules to conduct multi-relation reasoning link prediction rather than superficial vector triangle linkage in embedding models. By this way, we can explore relation between two entities in deeper context to achieve higher accuracy. In experiments, we demonstrate that EM-RBR achieves better performance compared with previous models on FB15k, WN18 and our new dataset FB15k-R, especially the new dataset where our model perform futher better than those state-of-the-arts. We make the implementation of EM-RBR available at \url{https://github.com/1173710224/link-prediction-with-rule-based-reasoning}.
\end{abstract}

\section{Introduction}
\label{sec:intro}
Knowledge graph (KG) has the ability to convey knowledge about the world and express the knowledge in a structured representation. The rich structured information provided by knowledge graphs has become extremely useful resources for many Artificial Intelligence related applications like query expansion~\citep{inproceedings}, word sense disambiguation~\citep{wasserman-pritsker-etal-2015-learning}, information extraction~\citep{10.5555/2002472.2002541}, etc. A typical knowledge representation in KG is multi-relational data, stored in RDF format, e.g. \emph{(Paris, Capital-Of, France)}. However, due to the discrete nature of the logic facts~\citep{wang2016learning}, the knowledge contained in the KG is meant to be incomplete~\citep{sadeghian2019drum}. Consequently, knowledge graph completion(KGC) has received more and more attention, which attempts to predict whether a new triplet is likely to belong to the knowledge graph (KG) by leveraging existing triplets of the KG.

Currently, the popular embedding-based KGC methods aim at embedding entities and relations in knowledge graph to a low-dimensional latent feature space. The implicit relationships between entities can be inferred by comparing their representations in this vector space. These researchers~\citep{bordes2013translating,mikolov2013distributed,wang2014knowledge,ji2015knowledge,lin2015learning,nguyen2017novel} make their own contributions for more reasonable and competent embedding. But the overall effect is highly correlated with the density of the knowledge graph. Because embedding method always fails to predict weak and hidden relations which a low frequency. The embedding will converge to a solution that is not suitable for triplets owned weak relations, since the training set for embedding cannot contain all factual triplets.However, reasoning over the hidden relations can covert the testing target to a easier one. For example, there is an existing triplet \emph{(Paul, Leader-Of, SoccerTeam)} and a rule \emph{Leader-Of(x,y) $\Longrightarrow$ Member-Of(x,y)} which indicates the leader of a soccer team is also a member of a sport team. Then we can apply the rule on the triplet to obtain a new triplet \emph{(Paul, Member-of, SportTeam)} even if the relation \emph{Member-of} is weak in knowledge graph.

Besides, some innovative models try to harness rules for better prediction. Joint models~\citep{rocktaschel2015injecting,wang2019logic,guo2016jointly} utilize the rules in loss functions of translation models and get a better embedding representation of entities and relations. An optimization based on ProPPR~\citep{wang2016learning} embeds rules and then uses those embedding results to calculate the hyper-parameters of ProPPR.
These efforts all end up on getting better embedding from rules and triplets, rather than solving completion through real rule-based reasoning, which is necessary to address weak relation prediction as mentioned before. Compared with them, EM-RBR can perform completion from the reasoning perspective.

% \emph{Challenge} 
We propose a novel framework EM-RBR combing embedding and rule-based reasoning, which is a BFS essentially. In the development of the joint framework EM-RBR, we meet two challenges. On the one hand, we use AMIE~\citep{galarraga2013amie} to auto-mine large amount of rules but not manually. However, these rules automatically mined sometimes are not completely credible. Therefore, it is necessary to propose a reasonable way to measure rules to pick proper rules when reasoning. On the other hand, it is known that traditional reasoning-based methods will give only 0 or 1 to one triplet to indicate acceptance or rejection for the given knowledge graph. This conventional qualitative analysis lacks the quantitative information as the embedding models. So the result of EM-RBR need to reflect the probability one triplet belonging to the knowledge graph.

% We propose a novel framework EM-RBR combing embedding and rule-based reasoning. Given a new triplet, we apply rules to conduct reasoning. With the help of the structured information of logic rules in the the same semantic space with embedding, a score can be output by EM-RBR to measure the quality of the triplet.

Three main contributions in EM-RBR are summarized as follows: 
\begin{itemize}
    \item EM-RBR is flexible and general enough to be combined with a lot of embedding models.
    \item We propose novel rating mechanism for triplets combined with reasoning process, which can distinguish a given triplet with other wrong triplets better.
    \item We propose a novel rating mechanism for auto-mined reasoning rules and each rule will be measured properly in our framework.
\end{itemize}

In the remaining of this paper, we will explain how our model works in Section~\ref{sec:method}, experiments in Section~\ref{sec:exp} and related work in Section~\ref{sec:rel}.

\section{Method}
\label{sec:method}
The core idea of our framework is to conduct multi-relation path prediction in deeper context from reasoning perspective, that is in the form of breadth first search. Before explaining the concrete reasoning algorithm, let's take an overview of our framework in Section~\ref{sec:overview}.

\begin{figure}[t]
    \centering
    \includegraphics[width=\linewidth]{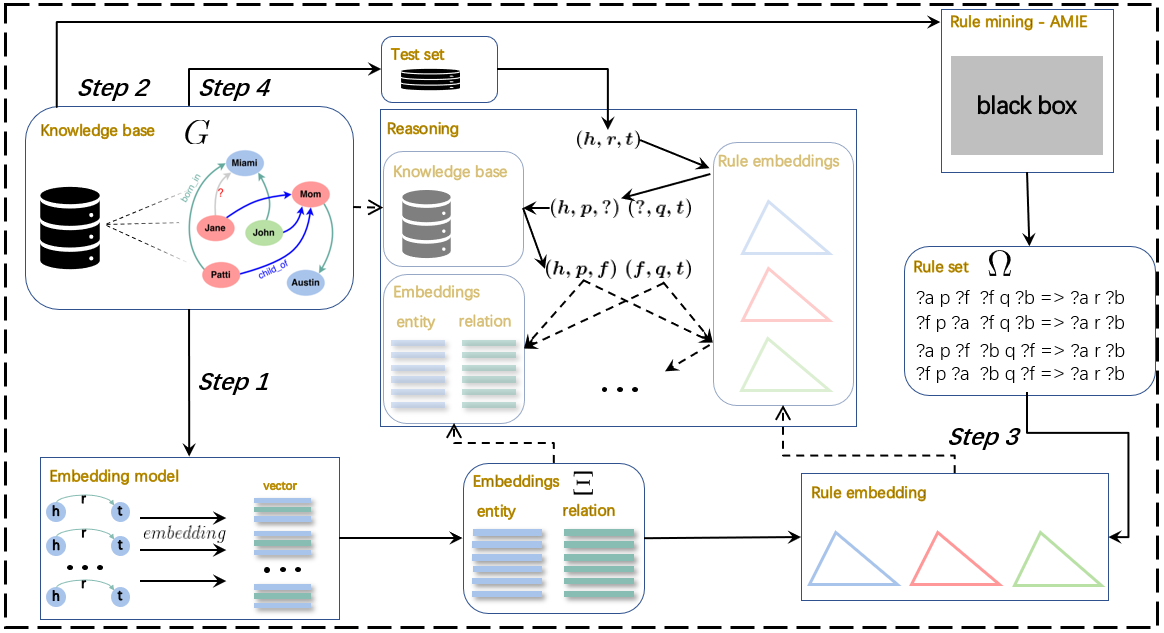}
    \caption{An overview of our framework.}
    \label{fig:flow}
\end{figure}

\subsection{Overview}
\label{sec:overview}
\begin{defn} \textbf{Rule:} A rule in our framework is in the form of $B_1(x,z) \wedge B_2(z,y) \Longrightarrow H(x,y)$ or $B(x,y) \Longrightarrow H(x,y)$, where the entities order in one triplet is random, i.e. $B_3(z,x) \wedge B_4(z,y) \Longrightarrow R(x,y)$ is also a valid rule. 
\label{def:rule}
\end{defn}

We model a knowledge graph as a collection of facts $G = \{(h,r,t)|h,t\in\mathcal{E},r\in\mathcal{R}\}$, where $\mathcal{E}$ and $\mathcal{R}$ represent the set of entities and relations in the knowledge graph, respectively. The steps of our framework are as follows corresponding to Figure~\ref{fig:flow}.
\begin{description}
    \item{\emph{\textbf{Step 1}}}. We invoke an embedding model to get a set $\Xi \in \mathbb{R}^{(|\mathcal{E}| + |\mathcal{R}|) \times k}$ containing the $k$-dimensional embedding of entities and relations in $G$.
    \item{\emph{\textbf{Step 2}}}. We apply AMIE~\citep{galarraga2013amie} on $G$ to get the reasoning rule set $\Omega$, where each rule meets Definition~\ref{def:rule}.
    \item{\emph{\textbf{Step 3}}}. The reasoning rules are measured based on the embedding of relations contained in the rule, which will explained in Secion~\ref{sec:handl}.
    \item{\emph{\textbf{Step 4}}}. Reasoning is conducted for a given triplet $(h,r,t)$, which will be described in Section~\ref{sec:process}.
\end{description}

\subsection{Reasoning algorithm}
\label{sec:process}
\begin{defn}\textbf{score:} The score $\Phi$ of a triplet meets $\Phi \geq 1$. The smaller $\Phi$ is, the triplet belongs to knowledge graph with greater probability. A triplet $(h,r,t)$'s score is denoted as $\Phi_{\sim(h,r,t)}$.
\label{def:score}
\end{defn}
We define two scores for each state during the search process. $\mathcal{H}$ is a heuristic score and $\mathcal{L}$ is the state score which will be used to compute $\Phi$. Our method is based on the idea of BFS. We use a priority queue $Q$ to store states in ascending order of $\mathcal{H}$. The initial state is the target triplet itself, whose $\mathcal{H}$ is 1 and $\mathcal{L}$ is its score under an embedding model. Push the initial state into $Q$.

During the search process, pop the top of $Q$ as the current state $s_{cur}$. It will not be extended if $\mathcal{H}_{s_{cur}}\geq \Phi$\footnote{$\Phi$ is defined as Definition~\ref{def:score} and initialized as $\mathcal{L}_{s_0}$. We will always update $\Phi$ after we pop a state from $Q$.}, otherwise we will extend it by matching rules to get new states. For each new state $s_{new}$, compute its score $\mathcal{H}_{s_{new}}$ and $\mathcal{L}_{s_{new}}$. If $\mathcal{H}_{s_{new}} < \mathcal{L}_{s_{cur}}$, the state will then be pushed into $Q$.

Repeat the above process until $Q$ is empty. Finally, we select the minimum $\mathcal{L}$ of all states that were pop from $Q$ as $\Phi$. The pseudo code is as shown in Appendix~\ref{sec:pcode}.

In the above procedure, we abstract three things: \emph{1.matching and extension}, \emph{2.state}, \emph{3.computation of $\mathcal{H},\mathcal{L}$}. Details are shown in the following subsections.

\subsubsection{matching and extension}
State is a set of triplets, the initial state is the target triplet itself. Intermediate states are extended from the initial state. So essentially, the extension of a state is extension of the triplets in the state. For a triplet $(h,r,t)$, the process of matching and extension is roughly as follows:
\begin{description}
    \item{\emph{\textbf{1.}}} Find rules $\omega \in \Omega$ in the form of $B_1(x,z) \wedge B_2(z,y) \Longrightarrow H(x,y)$\footnote{The rules we analyzed here are in the form of $B_1(x,z) \wedge B_2(z,y) \Longrightarrow H(x,y)$. As for rules like $B(x,z) \Longrightarrow H(x,y)$, the process is similar and will not be overtalked here.}, where $H = r$.
    \item{\emph{\textbf{2.}}} Assign entities to variables in the rule, i.e. $x = h,y = t$.
    \item{\emph{\textbf{3.}}} Find all $z_0$ that satisfy $(x,B_1,z_0) \in G$ or $(z_0,B_2,y) \in G$, where $x = h,y = t$.
    \item{\emph{\textbf{4.}}} $(h,r,t)$ is extended to $\{(h,B_1,z_0),(z_0,B_2,t)\}$. A triplet always has multiple extensions.    
\end{description}
For example, we expand the target triplet in the initial state. There are two triplets in the sub-state, and either of them must be in the knowledge graph. When the sub-state is further expanded, the triplet in the knowledge graph need not to be expanded. Therefore, there should be $m+1$ triplets in each sub-state after extending $m$ times. And at least $m$ of them belong to the knowledge graph.

\subsubsection{computation of $\mathcal{H}$ and $\mathcal{L}$}
\label{sec:handl}
$\mathcal{H}_{\sim\mathcal{O}} (h,r,t)$ denotes the heuristic score of triplet $(h,r,t)$ when extended to  state $\mathcal{O}$ and $\mathcal{L}_{\sim\mathcal{O}} (h,r,t)$ is the corresponding state score.
\begin{equation}
\label{equ:h}
\mathcal{H}_{\sim\mathcal{O}} (h,r,t) = \prod_{(B_1 \wedge B_2 \Rightarrow H) \in \Delta_{Path}} \omega(B_1,B_2,H)\ \ \ \boldsymbol{w.r.t},\ \omega(B_1,B_2,H) \gets e^{\frac{||\boldsymbol{B_1} + \boldsymbol{B_2} - \boldsymbol{H}||}{k}}
\end{equation}
$\mathcal{H}_{\sim\mathcal{O}} (h,r,t)$ is defined as Equation~\ref{equ:h} indicating the product of the scores of all the rules. $\Delta_{Path}$ represents the set of the rules used in the extension from the initial state to the current state. $\omega(B_1,B_2,H)$ is the score of rules in the shape of $B_1 \wedge B_2 \Rightarrow H$. 
\begin{equation}
\label{equ:l}
\mathcal{L}_{\sim\mathcal{O}} (h,r,t) = \mathcal{H}_{\sim\mathcal{O}} (h,r,t) * \prod_{(\mathcal{O}_h,\mathcal{O}_r,\mathcal{O}_t) \in \mathcal{O}} s_{\sim transX}(\mathcal{O}_h,\mathcal{O}_r,\mathcal{O}_t)
\end{equation}
$\mathcal{L}_{\sim\mathcal{O}} (h,r,t)$ is defined as Equation~\ref{equ:l} indicating the product of $\mathcal{H}_{\sim\mathcal{O}} (h,r,t)$ and the scores of all the triplets in the state. $\mathcal{O}$ denotes the state and $(\mathcal{O}_h,\mathcal{O}_r,\mathcal{O}_t)$ is a triplet belongs to $\mathcal{O}$. $s_{\sim transX}(\mathcal{O}_h,\mathcal{O}_r,\mathcal{O}_t)$ is the embedding score of this triplet as defined in Equation~\ref{equ:trans}.
\begin{equation}
\label{equ:trans}
s_{\sim transX}(\mathcal{O}_h,\mathcal{O}_r,\mathcal{O}_t) = \left\{
\begin{matrix}
&1\ &if\ (\mathcal{O}_h,\mathcal{O}_r,\mathcal{O}_t) \in G\\
&||\boldsymbol{\mathcal{O}_h}+\boldsymbol{\mathcal{O}_r}-\boldsymbol{\mathcal{O}_t}||/k + 1\ &if\ (\mathcal{O}_h,\mathcal{O}_r,\mathcal{O}_t) \notin G
\end{matrix}
\right.
\end{equation}

\emph{Rule's score} \\
To evaluate the score of rule $B_1(x,z) \wedge B_2(z,y) \Longrightarrow H(x,y)$, we visualize the three triplets of this rule in a two-dimensional space in Figure~\ref{fig:rule} of Appendix~\ref{sec:rule}. In our model, if a rule has a high confidence, it should satisfy   $\|\boldsymbol{x} + \boldsymbol{H} - \boldsymbol{y}\| \approx \|\boldsymbol{x} + \boldsymbol{B_1} - \boldsymbol{z} + \boldsymbol{z} + \boldsymbol{B_2} - \boldsymbol{y}\|$. We have $\boldsymbol{H} \approx \boldsymbol{B_1} + \boldsymbol{B_2}$, so we can use $\|\boldsymbol{B_1} + \boldsymbol{B_2} - \boldsymbol{H}\| \in \mathbb{R}^{3 \times k}$ to evaluate the score of the rule. $k$ is the dimension of embedding. The smaller score, the higher confidence. To make the dimension in the calculation uniform, we divide the score of the rule by $k$. And then perform the $e$ exponential transformation to get the form in the Equation~\ref{equ:h}. The reason for this transformation will be explained in section~\ref{sec:ana}.

\emph{Triplet's score}\\
$||\boldsymbol{\mathcal{O}_h}+\boldsymbol{\mathcal{O}_r}-\boldsymbol{\mathcal{O}_t}||$ is the score of triplet $(\mathcal{O}_h ,\mathcal{O}_r,\mathcal{O}_t)$ in transE model\footnote{Here we take transE as an example, so we use $||\boldsymbol{\mathcal{O}_h}+\boldsymbol{\mathcal{O}_r}-\boldsymbol{\mathcal{O}_t}||$. If the combined model changes, this formula should change to the form in the combined model, too.}. The smaller the value, the more likely the triplet is in $G$. When $(\mathcal{O}_h,\mathcal{O}_r,\mathcal{O}_t) \in G$, the score is assumed to be 0. The same to rule's score, we also perform a certain transformation on the scores of the triplets, which is to divide by $k$ and add 1.

\subsection{Example}
\label{sec:example}
\begin{assumption}
\label{assump:example}
We put all the necessary message in Table~\ref{tab:stateinexample} and~\ref{tab:ruleinexample}. Apart from that, we make two assumptions. One is that we use the same symbol $r_i$ to represent a rule's symbol and rule's score. Another is that we define some data relations as Equation~\ref{equ:datarela}.
\begin{equation}
\label{equ:datarela}
r_1 r_3 r_5 > \mathcal{L}_{\sim{s_3}} (h,r,t) \And \mathcal{H}_{\sim{s_7}} (h,r,t) > \Phi_{\sim(h,r,t)}
\end{equation}
\end{assumption}
\begin{figure}[h]
\begin{center}
\includegraphics[width=\linewidth]{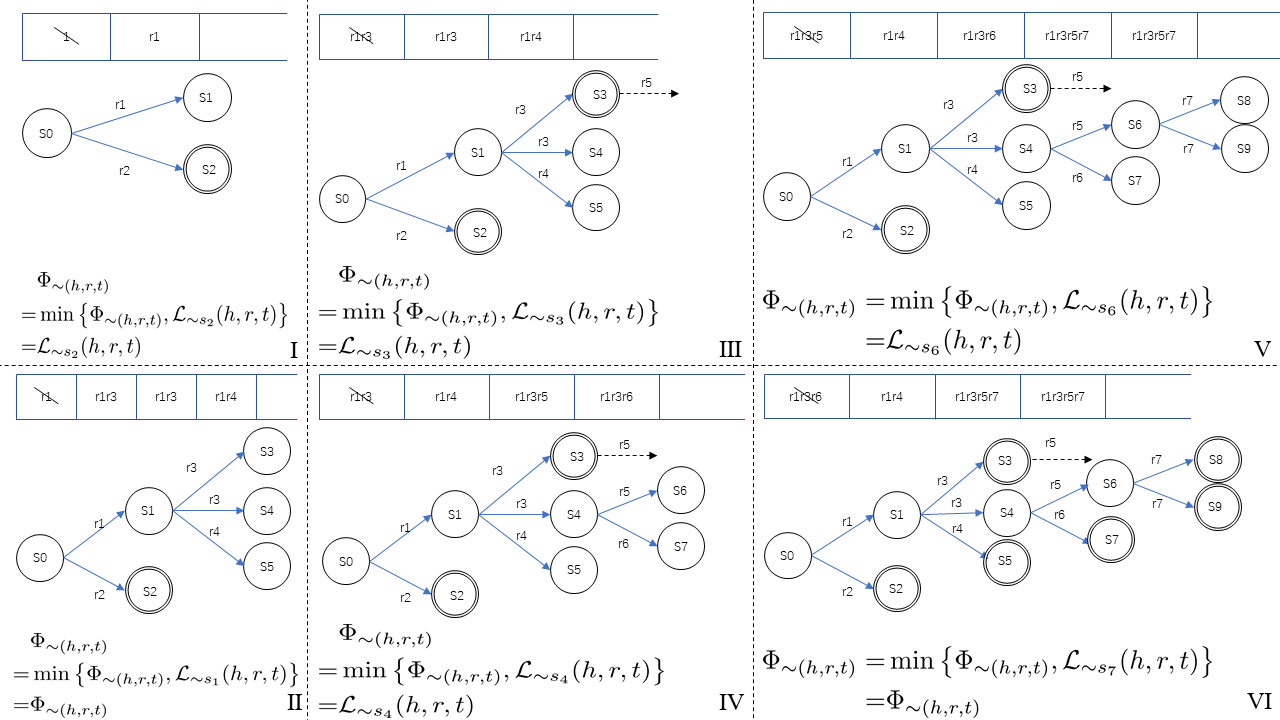}
\end{center}
\caption{Demonstration of the search process based on an example. The search process is divided into six stages, each stage is contained in a sub-graph, each sub-graph contains three parts. The top of the sub-graph shows the current state of the priority queue, the middle part is the visualization of the search, the formula for updating $\Phi_{\sim(h,r,t)}$ at each stage is given at the bottom.}
\label{fig:example}
\end{figure}
In this section, we use an example to illustrate our algorithm as shown in  Figure~\ref{fig:example}\footnote{There will be some conflicts in the usage of symbols. For these symbols, it's only valid in this example.}. The initial state $s_0$ only contains one triplet $(h,r,t)$, and its state score and heuristic score are both 1. At the beginning, the priority queue $Q$ has only one element, i.e. the initial state with its scores. The search process is as follows, and the necessary message is defined in Assumption~\ref{assump:example}.

\begin{itemize}
    \item[\Rmnum{1}.] $s_0$ matches $r_1$ and $r_2$ and extends to $s_1$ and $s_2$ respectively. $s_2$ is a termination state for the triplets in $s_2$ are all in $G$. We use $\mathcal{L}_{\sim s_2}(h,r,t)$ to update $\Phi_{\sim(h,r,t)}$ and push $s_1$ into $Q$.
    \item[\Rmnum{2}.] Pop the top of queue $s_1$. Use it to update $\Phi_{\sim(h,r,t)}$ and then extend it to three new states which will be pushed to $Q$.
    \item[\Rmnum{3}.] Pop the top of queue $s_3$ to update $\Phi_{\sim(h,r,t)}$ and extend it with matching the rule $r_5$. Since $r_1 r_3 r_5 > \mathcal{L}_{\sim{s_3}} (h,r,t)$, i.e. the solution produced by this path will not be the global minimum. As a consequence, this state is no longer extended.
    \item[\Rmnum{4}.] Pop the top of queue $s_4$ to update $\Phi_{\sim(h,r,t)}$ and extend to get two new states $s_6,s_7$.
    \item[\Rmnum{5}.] Pop the top of queue $s_6$ to update $\Phi_{\sim(h,r,t)}$ and extend to $s_8,s_9$ after the rule $r_7$.
    \item[\Rmnum{6}.] Pop the top of queue $s_7$ and now $\mathcal{H}_{\sim{s_7}} (h,r,t) > \Phi_{\sim(h,r,t) }$. So $s_7$ and the remaining states in $Q$ need not extend. Therefore, all the remaining states in $Q$ become termination states. The search stops.
\end{itemize}

\subsection{Analyse}
\label{sec:ana}
\subsubsection*{Is the algorithm sure to be effective?}
For three real number $a,b,c(a,b,c>1)$, it's possible that $c > a * b * 1$. Consider triplet $(h,r,t)$ and rule $B_1(x,z) \wedge B_2(z,y) \Longrightarrow H(x,y)$ ,w.r.t $r=H$. If $c$ represents the score of $(h,r,t)$, $a$ represents the score of the rule, $b$ represents the score of the expanded new triplet that not in the knowledge graph, and $1$ represents the score of the expanded new triplet that in the knowledge graph. Then the score of $(h,r,t)$ will be reduced to $a * b$, i.e. we use the new expanded triplets and the rule to evaluate $(h,r,t)$.

Of course, this optimization will be effective only on the correct triplets. For the wrong triplets, another wrong triplet with a large score will be obtained after rule matching. So $c > a * b$ in this occasion is a very unlikely event. As a result, the correct triplets are optimized, and the wrong triplets will generally not be optimized. Therefore, from a macro perspective, the ranking of the correct triplets will increase.

\subsubsection*{Is this algorithm a terminating algorithm?}
The heuristic score of a state is the product of scores of all the rules along the reasoning path. The scores of the rules are all number greater than 1, so when a search path is long enough, $\mathcal{H}$ must be greater than $\Phi$. So the search must be able to stop.
\subsubsection*{What is the principle when designing the calculation of rules and triplets?}
From the above, we require that the scores of rules and triplets are greater than 1 and close to 1. Given that each dimension of embedding obtained by the translation model is a number less than 1 and close to 0, we divide the score by the corresponding dimension $k$ and then add 1 to meet our requirements. In addition, in order to highlight the importance of rules in the calculation, we use exponential changes for rules instead of plus 1.
\section{Experiment}
\label{sec:exp}
\subsection{Experiment setup}
\textbf{Dataset:} We evaluate EM-RBR on FB15k, WN18~\citep{bordes2013translating} and FB15k-R. FB15k has 14951 entities, 1345 relations and 59071 test triplets in total. WN18 has 40943 entities, 18 relations and 5000 test triplets in total. We create FB15k-R, a subset of FB15k, which contains 1000 tested triplets that have rich rules to take reasoning.

\textbf{Metrics:} We use a number of commonly used metrics, including Mean Rank (MR), Mean Reciprocal Rank (MRR), and Hit ratio with cut-off values n = 1,10. MR measures the average rank of all correct entities and MRR is the average inverse rank for correct entities. Hits@n measures the proportion of correct entities in the top n entities. MR is always greater or equal to 1 and the lower MR indicates better performance, while MRR and Hits@n scores always range from 0.0 to 1.0 and higher score reflects better prediction results. We use filtered setting protocol~\citep{bordes2013translating}, i.e.,  filtering out any corrupted triples that appear in the KB to avoid possibly flawed evaluation.

\textbf{Baseline:} To demonstrate the effectiveness of EM-RBR, we compare with a number of competitive baselines: TransE~\citep{bordes2013translating}, TransH~\citep{wang2014knowledge}, TransR~\citep{lin2015learning}, TransD~\citep{ji2015knowledge}, RUGE~\citep{guo2017knowledge}, ComplEx~\citep{DBLP:journals/corr/TrouillonWRGB16} and DistMult~\citep{kadlec2017knowledge}. Among these state-of-arts, TransE, TransH, TransR and TransD are combined with our reasoning framework. These 8 models are evaluated on FB15k and WN18 to prove that our framework is a real reinforced framework. In the end, all the baselines and combined models are evaluated on FB15k-R. 

\textbf{Implementation:} For TransE, TransH, TransR and TransD, we set the same parameters, i.e., the dimensions of embedding $k = 100$, learning rate $\lambda = 0.001$, the margin $\gamma = 1$. We traverse all the training triplets
for 1000 rounds. Other parameters of models are set as the same with the parameters in the published works ~\citep{bordes2013translating,wang2014knowledge,lin2015learning}\footnote{We use the implementation of these model at \url{https://github.com/thunlp/Fast-TransX}}. For RUGE, we set the embedding dimension $k = 100$ and other hyper-parameters are the same with ~\citet{guo2017knowledge}\footnote{The code is available at \url{https://github.com/iieir-km/RUGE}}. For ComplEx and DistMult, all the parameters are consistent with~\citet{DBLP:journals/corr/TrouillonWRGB16}\footnote{\url{https://github.com/ttrouill/complex}}.

\subsection{Experiment results}
\begin{table}[t]
\centering
\small
\caption{Experimental results on FB15k,WN18 and FB15k-R test set. [$\boldsymbol{\ddagger}$]:E-R(E) denotes EM-RBR(E), indicating that the embedding model in this experimental group is transE. [$\boldsymbol{\star}$]:We don't use it here because it's time-consuming and not better than transE on WN18 as reported in ~\citet{lin2017learning}. [$\boldsymbol{\dotplus}$]:This model is rerun and only tested on the subset  of FB15k to show that our embedding model can perform better than it.}

\begin{tabular}{C{0.8cm}
L{0.55cm}L{0.55cm}L{0.55cm}L{0.55cm}C{-2cm}
L{0.55cm}L{0.55cm}L{0.55cm}L{0.55cm}C{-2cm}
L{0.55cm}L{0.55cm}L{0.55cm}L{0.55cm}}
\toprule
\multirow{2}{*}{\textbf{Model}} & \multicolumn{4}{c}{\textbf{FB15k}} && \multicolumn{4}{c}{\textbf{WN18}} && \multicolumn{4}{c}{\textbf{FB15k-R}}\\
\cline{2-5}\cline{7-10}\cline{12-15}
  & MR & MRR & H@1 & H@10 & & MR & MRR & H@1 & H@10 & & MR & MRR & H@1 & H@10\\
%   & MR & MRR & H1 & H3 & H5 & H10 && MR & MRR & H1 & H3 & H5 & H10\\
  \midrule
TransE&70.3&45.77&29.98&74.27&&200.9&57.47&23.21&97.68&&71.33&26.11&14.9&48.1\\\noalign{\smallskip}
TransH&72.56&45.81&30.37&74.01&&210.7&61.94&32.03&97.49&&50.65&30.43&18.25&54.95\\\noalign{\smallskip}
TransR&55.98&47.88&31.1&77.04&&{\bf $\boldsymbol{\star}$\ - -}&{\bf \ \ - -}&{\bf \ \ - -}&{\bf \ \ - -}&&29.64&18.51&7.65&37.6\\\noalign{\smallskip}
TransD&56.41&47.88&32.48&75.99&&202.8&60.35&29.6&97.37&&28.24&26.16&14&50.45\\\noalign{\smallskip}
\midrule
\tabincell{c}{$\boldsymbol{\ddagger}$E-R(E)}&68.36&50.01&34.44&76.23&&\bf198.1&\bf85.23&73.94&\bf97.83&&3.12&79.88&65.1&96.4\\\noalign{\smallskip}
\tabincell{c}{E-R(H)}&70.72&52.39&\bf38.82&76.52&&201.4&84.57&74.97&96.48&&3.52&85.61&\bf75.45&97.8\\\noalign{\smallskip}
\tabincell{c}{E-R(R)}&55.47&51.93&35.86&\bf78.35&&{\bf $\boldsymbol{\star}$\ - -}&{\bf \ \ - -}&{\bf \ \ - -}&{\bf \ \ - -}&&\bf1.73&\bf86.01&74.3&\bf99.2\\\noalign{\smallskip}
\tabincell{c}{E-R(D)}&\bf55.21&\bf53.02&38.25&78.33&&201.8&84.63&\bf75.21&97.5&&3.245&82.04&70.45&96.15\\\noalign{\smallskip}
\midrule

$\boldsymbol{\dotplus}$RotateE&{\bf \ \ - -}&{\bf \ \ - -}&{\bf \ \ - -}&{\bf \ \ - -}&&{\bf \ \ - -}&{\bf \ \ - -}&{\bf \ \ - -}&{\bf \ \ - -}&&\underline{\emph{26.66}}&33.29&20.05&59.75\\\noalign{\smallskip}

RUGE&{\bf \ \ - -}&{\bf \ \ - -}&{\bf \ \ - -}&{\bf \ \ - -}&&{\bf \ \ - -}&{\bf \ \ - -}&{\bf \ \ - -}&{\bf \ \ - -}&&53.63&49.14&33.05&78.2\\\noalign{\smallskip}

ComplEx&{\bf \ \ - -}&{\bf \ \ - -}&{\bf \ \ - -}&{\bf \ \ - -}&&{\bf \ \ - -}&{\bf \ \ - -}&{\bf \ \ - -}&{\bf \ \ - -}&&51.43&\underline{\emph{51.1}}&\underline{\emph{35.8}}&\underline{\emph{79.0}}\\\noalign{\smallskip}

DistMult&{\bf \ \ - -}&{\bf \ \ - -}&{\bf \ \ - -}&{\bf \ \ - -}&&{\bf \ \ - -}&{\bf \ \ - -}&{\bf \ \ - -}&{\bf \ \ - -}&&62.02&46.2&30.2&77.1\\\noalign{\smallskip}
\bottomrule
\end{tabular}
\label{tab:result}
\end{table}
Experimental Results are as shown in Table~\ref{tab:result}. Through this experiment, we would like to prove two things. One is that EM-RBR is a valid reinforced model, i.e. EM-RBR(X) model always performs better than X model. Another is that EM-RBR will beat all of the current state-of-the-arts on a data set with rich rules.

When evaluating on FB15k and WN18, our model has improved all the metrics compared with the translation model in the baseline, especially MRR and Hits@1 on WN18. For example, EM-RBR(D) improve Hits@1 on WN18 from \emph{0.296} to \emph{0.752} compared to transD.

As for FB15k-R, each triplet in this data set can match a lot of rules so that they can be optimized extremely under EM-RBR. On this data set, the best MR is \emph{1.73} from EM-RBR(R) which is an improvement of \emph{24.93} relative to RotateE. The best MRR is \emph{0.86} from EM-RBR(R) which is an improvement of \emph{0.35} relative to ComplEx. The best Hits@1 is \emph{0.7545} from EM-RBR(H) which is an improvement of \emph{39.65} relative to ComplEx. The best Hits@10 is \emph{0.992} from EM-RBR(R) which is an improvement of \emph{20.2} relative to ComplEx.

\subsection{Result analysis}
The triplets in the test set of each data set can be roughly divided into two parts: one is that the rules can be matched to be optimized by our model, and the other is that without rules to be matched. In FB15k, the ratio of these two parts is about 1:5. From the final result, although various metrics have been improved compared with the model before the combination. In fact, only a small part of the triplets have been optimized. It is also for this reason that the capabilities of our model can be fully demonstrated on the FB15k-R, because each triplet in this set has many rules that can be matched to obtain a good optimization effect.

In order to better understand the specific situation being optimized on each triplet. We respectively analyzed the corresponding ranking of each triplet under the translation model and the EM-RBR model when the head entity replacement and tail entity replacement were performed. The results were displayed in Table ~\ref{tab:analyse}. The data item in the table is the result of sorting from largest to smallest value of $s_{\sim trans}-s_{\sim ER}$, where $s_{\sim trans}$ is the ranking under the corresponding translation model and $s_{\sim E-R}$ is the ranking under the corresponding EM-RBR model.

\begin{table}[t]
\centering
\footnotesize
\caption{Optimized case analysis. [$\boldsymbol{\dotplus}$]: the id number of the test case, for example, the first test case is \emph{/m/01qscs	/award/award\_nominee/award\_nominations./award/award\_nomination/award	/m/02x8n1n} and its id number is 0. [$\boldsymbol{\ast}$]: the rank of the test case in EM-RBR. [$\boldsymbol{\ddagger}$]: the rank of the test case in the embedding model. [$\boldsymbol{\diamond}$]: L corresponds to replacing the head entity and R the tail entity.}
\begin{tabular}{
C{0.6cm}
C{0.6cm}C{0.7cm}C{0.5cm}C{0.5cm}
C{-10cm}
C{0.6cm}C{0.7cm}C{0.5cm}C{0.5cm}
C{-10cm}
C{0.6cm}C{0.7cm}C{0.5cm}C{0.5cm}}
\toprule
\multirow{2}{*}{\textbf{Rank}} & \multicolumn{4}{c}{EM-RBR(\textbf{E})} && \multicolumn{4}{c}{EM-RBR(\textbf{H})} && \multicolumn{4}{c}{EM-RBR(\textbf{R})}\\
\cline{2-5}\cline{7-10}\cline{12-15}
 &$\boldsymbol{\dotplus}$id & $\boldsymbol{\ast}$E-R & $\boldsymbol{\ddagger}$trans & $\boldsymbol{\diamond}$L/R && $\boldsymbol{\dotplus}$id & $\boldsymbol{\ast}$E-R & $\boldsymbol{\ddagger}$trans & $\boldsymbol{\diamond}$L/R && $\boldsymbol{\dotplus}$id & $\boldsymbol{\ast}$E-R & $\boldsymbol{\ddagger}$trans & $\boldsymbol{\diamond}$L/R\\
\midrule
1 & 47722 & 2 & 14141 & R && 18355 & 2 & 12689 & L && 15105 & 2 & 966 & L\\
2 & 47722 & 2 & 13900 & L && 32966 & 3 & 8551 & L && 42675 & 1 & 868 & R\\
3 & 18355 & 2 & 7525 & L && 18355 & 2 & 7231 & R && 34891 & 2 & 733 & R\\
4 & 36133 & 2 & 6884 & L && 47722 & 2 & 4569 & L && 24314 & 1 & 714 & R\\
5 & 33004 & 1 & 6253 & L && 24243 & 1 & 4547 & L && 32849 & 1 & 701 & L\\
6 & 33243 & 2 & 5883 & R && 33004 & 1 & 4490 & L && 55951 & 2 & 673 & L\\
7 & 30883 & 2 & 5674 & R && 47722 & 2 & 3977 & R && 38773 & 1 & 640 & L\\
8 & 14035 & 2 & 4862 & L && 13358 & 5 & 3741 & R && 54283 & 52 & 674 & L\\
9 & 18355 & 2 & 4525 & R && 55951 & 2 & 3699 & L && 25500 & 1 & 585 & R\\
10 & 24243 & 1 & 3655 & L && 50019 & 1 & 3386 & R && 34891 & 2 & 555 & L\\
$\boldsymbol{\cdot\ \cdot\ \cdot\ }$\\
19372 & 52886 & 4 & 2 & R && 23339 & 6 & 1 & L && 44273 & 2 & 1 & R\\
19373 & 52707 & 13 & 11 & L && 23288 & 7 & 2 & R && 43969 & 2 & 1 & R\\
19374 & 52529 & 9 & 7 & R && 23218 & 7 & 2 & R && 43664 & 2 & 1 & L\\
19375 & 51447 & 3 & 1 & R && 21906 & 7 & 2 & R && 43483 & 2 & 1 & L\\
19376 & 50932 & 4 & 2 & R && 20794 & 7 & 2 & R && 42380 & 2 & 1 & R\\
$\boldsymbol{\cdot\ \cdot\ \cdot\ }$\\
\bottomrule
\end{tabular}
\label{tab:analyse}
\end{table}

\section{Related work}
\label{sec:rel}
For the path-based methods, \citet{lao2011random} uses Path Ranking Algorithm (PRA)~\citep{journals/ml/LaoC10} to estimate the probability of an unseen triplet as a combination of weighted random walks.
\citet{zhang2020efficient} and \citep{qu2019probabilistic} are both the combination of Markov logic network and embedding.
\citet{kok2007statistical} is mainly a clustering algorithm, clustering entity sets under multiple relationship categories.
\citet{gardner2014incorporating} makes use of an external text corpus to increase the connectivity of KB. The Neural LP model ~\citep{yang2017differentiable} compiles inferential tasks into differentiable numerical matrix sequences.
Besides, many studies have modeled the path-finding problem as a Markov decision-making process, such as the DeepPath model ~\citep{DBLP:journals/corr/XiongHW17} and MINERVA ~\citep{DBLP:journals/corr/abs-1711-05851}. 
For the embedding methods, \citet{nguyen2017overview} has organized the existing work. Our paper divides all embedding methods into four categories, which are: translation, Bilinear$\And$Tensor, neural network and complex vector. Firstly, for translation, the Unstructured model~\citep{bordes2014semantic} assumes that the head and tail entity vectors are similar without distinguishing relation types. The Structured Embedding (SE) model~\citep{bordes2011learning} assumes that the head and tail entities are similar only in a relation-dependent subspace. Later, there are transE, transR, transH~\citep{lin2015learning,wang2014knowledge,bordes2013translating}, etc. 
\citet{sadeghian2019drum} mines first-order logical rules from knowledge graphs and uses those rules to solve KBC. Additionally, other work~\citep{yang2017differentiable,galarraga2013amie} can extract some high-quality rules from knowledge base. 
For the second type, DISTMULT~\citep{yang2014embedding} is based on the Bilinear model~\citep{nickel2011three} where each relation is represented by a diagonal matrix rather than a full matrix. SimplE~\citep{kazemi2018simple} extends CP models~\citep{hitchcock1927expression} to allow two embeddings of each entity to be learned dependently. The third method is to implement embedding with a neural network. Apart from the models mentioned in Section~\ref{sec:intro}, NTN~\citep{socher2013reasoning} and ER-MLP~\citep{dong2014knowledge} also belong to this method. Fourthly, instead of embedding entities and relations in  real-valued vector space, ComplEx ~\citep{DBLP:journals/corr/TrouillonWRGB16} is an extension of DISTMULT in the complex vector space. ComplEx-N3 ~\citep{Lacroix2018CanonicalTD} extends ComplEx with weighted nuclear 3-norm. Also in the complex vector space, RotatE ~\citep{DBLP:journals/corr/abs-1902-10197} defines
each relation as a rotation from the head entity to the tail entity. QuatE ~\citep{DBLP:journals/corr/abs-1904-10281} represents entities by quaternion embeddings (i.e., hypercomplex-valued embeddings)
and models relations as rotations in the quaternion space.

\section{Conclusion $\And$ Future work}
\label{sec:con}
This paper introduces an innovative framework called EM-RBR combining embedding and rule-based reasoning, which can be easily integrated with any translation based  embedding model. Unlike previous joint models trying to get better embedding results from rules and triplets, our model allows solving completion from the reasoning perspective by conducting multi-relation path prediction, i.e. a breadth first search. We also demonstrate that EM-RBR can efficiently improve the performance of embedding methods for KGC. This makes the existing translation based embedding methods more suitable and reliable to be used in the real and large scale knowledge inference tasks.

There are two possible directions in the future. On one hand, we will combine our model with more embedding models, not just the translation-based embedding model. On the other hand, we are going to extract more and more reliable association rules to optimize our work. As mentioned above, only a part of triples are optimized when evaluating on FB15k. The fundamental reason for the rest is that there is no corresponding rule for matching. If these two problems are solved, EM-RBR can be better improved.

\newpage
\bibliography{iclr2021_conference}
\bibliographystyle{iclr2021_conference}
\newpage
\appendix
\section{Rule's visualization}
\label{sec:rule}
\begin{figure}[h]
    \centering
    \includegraphics[width=0.3\linewidth]{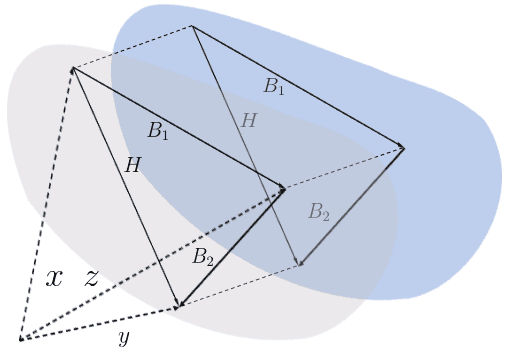}
    \caption{visualization of a rule.}
    \label{fig:rule}
\end{figure}

\section{Pseudo code}
\label{sec:pcode}
\begin{algorithm}[h]
	\caption{EM-RBR}
		\label{algo:code}
		\centering
	    \begin{algorithmic}[1]
		    \Require $(h,r,t)$
		    \Ensure $\Phi_{\sim(h,r,t)}$
		    \State Initialize $Q$ as an empty priority queue
		    \State Initialize the first state: $s_0 \gets \{(h,r,t)\}$, $\mathcal{H}_{\sim s_0} (h,r,t),\mathcal{L}_{\sim s_0} (h,r,t) \gets 1,1$
            \State Initialize the score: $\Phi_{\sim(h,r,t)} \gets (s_{\sim transX}(h,r,t))/k + 1$
		    \State $Q$.push($s_0$)
		    \While {\textbf{!}$Q$.empty()}
		        \State $s_{cur} \gets Q$.pop()
		        \State $\Phi_{\sim(h,r,t)} \gets \min\{\Phi_{\sim(h,r,t)},\mathcal{L}_{\sim s_{cur}} (h,r,t)\}$
		        \If {$\mathcal{H}_{\sim s_{cur}} (h,r,t) < \Phi_{\sim(h,r,t)}$}
                \State $\mathbb{S}_{ne} \gets$ extend($s_{cur}$)
                \State calculate($\mathcal{H}_{\sim s_{ne}} (h,r,t)$), $s_{ne} \in \mathbb{S}_{ne}$
                \State calculate($\mathcal{L}_{\sim s_{ne}} (h,r,t)$), $s_{ne} \in \mathbb{S}_{ne}$
                \For {$s_{ne} \in \mathbb{S}_{ne}$}
	            \If {$\mathcal{H}_{\sim s_{ne}} (h,r,t) < \mathcal{L}_{\sim s_{cur}} (h,r,t)$}
	            \State $Q$.push($s_{ne}$)
	            \EndIf
	            \EndFor
	            \EndIf
		    \EndWhile
        \end{algorithmic}
\end{algorithm}

\section{Data message in example}

\begin{table}[h]
\centering
\caption{Triplets in each state. [$\boldsymbol{\star}$]: If this symbol appears in the upper right corner of a triple, the triplet is not in the knowledge graph. Other triplets are all in the knowledge graph.}
\begin{tabular}{L{0.1\linewidth}L{0.14\linewidth}L{0.14\linewidth}L{0.14\linewidth}L{0.14\linewidth}L{0.14\linewidth}}
\toprule
\textbf{state}&\textbf{triplets}&&&&\\
\midrule
$s_1$ & $(h,B_1,m_1)^{\boldsymbol{\star}}$&$(m_1,B_2,t)$\\
$s_2$ & $(h,B_3,m_2)$&$(m_2,B_4,t)$\\
$s_3$ & $(h,B_5,m_3)$&$(m_3,B_6,m_1)^{\boldsymbol{\star}}$&$(m_1,B_2,t)$\\
$s_4$ & $(h,B_5,m_4)^{\boldsymbol{\star}}$&$(m_4,B_6,m_1)$&$(m_1,B_2,t)$\\
$s_5$ & $(h,B_7,m_5)^{\boldsymbol{\star}}$&$(m_5,B_8,m_1)$&$(m_1,B_2,t)$\\
$s_6$ & $(h,B_9,m_6)$&$(m_6,B_{10},m_4)^{\boldsymbol{\star}}$&$(m_4,B_6,m_1)$&$(m_1,B_2,t)$\\
$s_7$ & $(h,B_{11},m_7)$&$(m_7,B_{12},m_4)^{\boldsymbol{\star}}$&$(m_4,B_6,m_1)$&$(m_1,B_2,t)$\\
$s_8$ & $(h,B_9,m_6)$&$(m_6,B_{13},m_8)^{\boldsymbol{\star}}$&$(m_8,B_{14},m_4)$&$(m_4,B_{6},m_1)$&$(m_1,B_{2},t)$\\
$s_9$ & $(h,B_9,m_6)$&$(m_6,B_{13},m_9)$&$(m_9,B_{14},m_4)^{\boldsymbol{\star}}$&$(m_4,B_{6},m_1)$&$(m_1,B_{2},t)$\\
% \midrule
\bottomrule
\end{tabular}
\label{tab:stateinexample}
\end{table}

\begin{table}[t]
\centering
\caption{Rule's score}
\begin{tabular}{L{0.066\linewidth}L{0.088\linewidth}L{0.15\linewidth}L{0.07\linewidth}}
\toprule
\textbf{rule}&&&\textbf{score}\\
\midrule
$B_{1}(x,z)$&$ \wedge B_{2}(z,y)$&$ \Rightarrow r(x,y)$ & r1\\
$B_{3}(x,z)$&$ \wedge B_{4}(z,y)$&$ \Rightarrow r(x,y)$ & r2\\
$B_{5}(x,z)$&$ \wedge B_{6}(z,y)$&$ \Rightarrow B_{1}(x,y)$ & r3\\
$B_{7}(x,z)$&$ \wedge B_{8}(z,y)$&$ \Rightarrow B_{1}(x,y)$ & r4\\
$B_{9}(x,z)$&$ \wedge B_{10}(z,y)$&$ \Rightarrow B_{5}(x,y)$ & r5\\
$B_{11}(x,z)$&$ \wedge B_{12}(z,y)$&$ \Rightarrow B_{5}(x,y)$ & r6\\
$B_{13}(x,z)$&$ \wedge B_{14}(z,y)$&$ \Rightarrow B_{10}(x,y)$ & r7\\
% \midrule
\bottomrule
\end{tabular}
\label{tab:ruleinexample}
\end{table}

\end{document}